# Image Processing Based Forest Fire Detection

Vipin V

*Asst. Professor, Department of ECE,*
*St.Joseph's College of Engineering, Palai, India*
`vipinvenu85@gmail.com`

*Abstract:* **A novel approach for forest fire detection using image processing technique is proposed. A rule based color model for fire pixel classification is used. The proposed algorithm uses RGB and $YC_bC_r$ color space. The advantage of using $YC_bC_r$ color space is that it can separate the luminance from the chrominance more effectively than RGB color space. The performance of the proposed algorithm is tested on two sets of images, one of which contains fire; the other contains fire-like regions. Standard methods are used for calculating the performance of the algorithm. The proposed method has both higher detection rate and lower false alarm rate. Since, the algorithm is cheap in computation it can be used for real time forest fire detection.**

## I. INTRODUCTION

Forest fires represent a constant threat to ecological systems, infrastructure and human lives. Past has witnessed multiple instances of forest and wild land fires. Fires play a remarkable role in determining landscape structure, pattern and eventually the species composition of ecosystems. The integral part of the ecological role of the forest fires is formed by the controlling factors like the plant community development, soil nutrient availability and biological diversity. Fires are considered as a significant environmental issue because they cause prominent economical and ecological damage despite endangering the human lives [1].

According to the prognoses, forest fire, including fire clearing in tropical rain forests, will halve the world forest stand by the year 2030 [2]. Every year in Europe over 10.000 $km^2$ of forest terrain is burnt and in Russia and USA over 100.000 $km^2$. The Global fires during the year 2008 for the months of August and February. The fact that more than 20% of complete world $CO_2$ emissions comes from forest fires indicates that it is a phenomenon which has to be dealt with great attention. Also due to the global warming the rate of occurrence of forest fire has increased.

Traditional fire protection methods use mechanical devices or humans to monitor the surroundings. The most frequently used fire smoke detection techniques are usually based on particle sampling, temperature sampling, and air transparency testing. An alarm is not raised unless the particles reach the sensors and activate them.

Some of the methods are mentioned below:-

*A. Fire Watch Tower*

In watch towers human are made to observe the location throughout. If any fire occurs he reports it. However, accurate human observation may be limited by operator fatigue, time of day, time of year, and geographic location.

*B. Wireless Sensor Networks*

In a wireless sensor-based fire detection system, coverage of large areas in forest is impractical due to the requirement of regular distribution of sensors in close proximity and also battery charge is a big challenge [3].

*C. Satellite and Aerial Monitoring*

Satellites based system can monitor a large area, but the resolution of satellite imagery is low [8]. A fire is detected when it has grown quite a lot, so real time detection cannot be provided. Moreover, these systems are very expensive [4]. Weather condition (e.g. clouds) will seriously decrease the accuracy of satellite-based forest fire detection as the limitations led by the long scanning period and low resolution of satellites [5].

The motivation for an image processing based approach is due to rapid growth of the electronics; digital camera technology has grown such that cheap CCD and CMOS digital cameras are available in market with decently good resolution. Most of these cameras can be directly connected to the computer and store the captured images to the computer. Computer-vision-based systems which utilize digital camera technology and image/video processing techniques play a very promising role to effectively replace conventional forest fire detection systems.

The fire detection performance depends critically on the performance of the flame pixel classifier which generates seed areas on which the rest of the system operates. The flame pixel classifier is thus required to have a very high detection rate and preferably a low false alarm rate. There exist few algorithms which directly deal with the flame pixel classification in the literature. The flame pixel classification can be considered both in grayscale and color images. Most of the works on flame pixel classification in color image or video sequences are rule based.

Krull et al. [6] used low-cost CCD cameras to detect fires in the cargo bay of long range passenger aircraft. The method uses statistical features, based on grayscale video frames, including mean pixel intensity, standard deviation, and second-order moments, along with non-image features such as humidity and temperature to detect fire in the cargo compartment. The system is commercially used in parallel to standard smoke detectors to reduce the false alarms caused by the smoke detectors. The system also provides visual inspection capability which helps the aircraft crew to confirm the presence or absence of fire. However, the statistical image features are not considered to be used as part of a standalone fire detection system.

Marbach et al. [10] used YUV color model for the representation of video data, where time derivative of luminance component Y was used to declare the candidate fire pixels and the Chrominance components U and V were used to classify the candidate pixels to be in the fire sector or not. In addition to luminance and chrominance they have incorporated motion into their work. They report that their algorithm detects less than one false alarm per week; however, they do not mention the number of tests conducted.

Celik et al. [12] used normalized RGB (rgb) values for a generic color model for the flame. The normalized RGB is proposed in order to alleviate the effects of changing illumination. The generic model is obtained using statistical analysis carried out in r–g, r–b, and g–b planes. Due to the distribution nature of the sample fire pixels in each plane, three lines are used to specify a triangular region representing the region of interest for the fire pixels. Therefore, triangular regions in respective r–g, r–b, and g–b planes are used to classify a pixel. A pixel is declared to be a fire pixel if it falls into three of the triangular regions in r–g, r–b, and g–b planes.

Celik et al. [13] proposed a novel model for detection of fire and smoke detection using image processing approach. For fire detection the proposed method uses RGB and YCbCr color space. Few rules are identified to fire pixels, and then given to a Fuzzy Inference System (FIS). A rule table is formed depending on the probability value the pixel is considered to be fire. They report to have 99% accuracy but, this cannot be used for real time monitoring. In case of smoke detection they have given some threshold values but, this method may fail because the texture of smoke varies depending on the materials which are burned.

## II. PROPOSED FOREST FIRE MONITORING SYSTEM

In the proposed method image captured by the digital camera as input, process the image and performs actions as shown in figure 1. In this paper we will discuss about how fire regions are detected in the captured image.

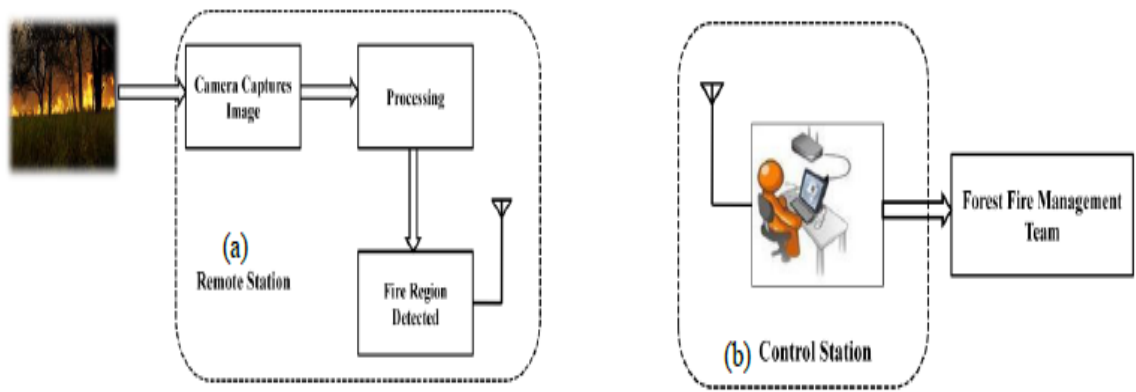

**Fig.1: Forest Fire monitoring System (a) Remote station and (b) Control Station.**

### III. CLASSIFICATION OF FIRE PIXEL

This section covers the detail of the proposed fire pixel classification algorithm. Figure 2 shows the flow chart of the proposed algorithm. Rule based color model approach has been followed due to its simplicity and effectiveness. For that, color space RGB and $YC_bC_r$ is chosen. For classification of a pixel to be fire we have identified seven rules. If a pixel satisfies these seven rules, we say that pixel belong to fire class.

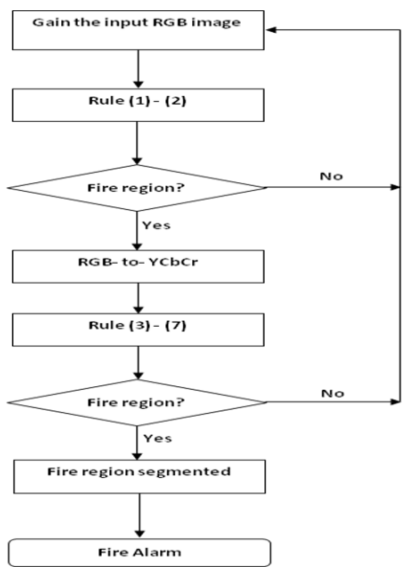

**Fig.2: Flow chart of proposed algorithm for forest fire detection.**

A digital colored image has three planes: Red, Green and Blue (R, G, and B). The combination of RGB color planes gives ability to devices to represent a color in digital environment. Each color plane is quantized into discrete levels. Generally 256 (8 bits per color plane) quantization levels are used for each plane, for instance white is represented by (R, G, B) = (255, 255, 255) and black is represented by (R, G, B) = (0, 0, 0). A color image consists of pixels, where each pixel is represented by spatial location in rectangular grid (x, y), and a color vector (R(x, y), G(x, y), B(x, y)) corresponding to spatial location (x, y).

#### A. Rule I

It can be noticed from figure 3 that for the fire regions, R channel has higher intensity values than the G channel, and G channel has higher intensity values than the B channel.

In order to explain this idea better, we picked sample images from figure 4 (a), and segmented its fire pixels as shown in figure 4 (b) with green color. Then we calculate mean values of R, G, and B planes in the segmented fire regions of the original images.

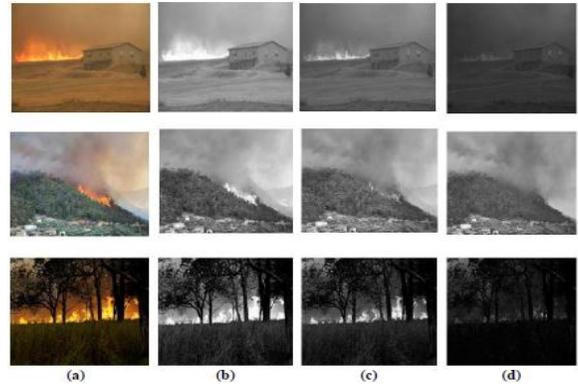

**Fig.3: Original RGB image [19] in column (a), and R, G, and B channels in column (b)-(d), respectively.**

The results are given in Table I for the images given in figure 4. It is clear that, on the average, the fire pixels show the characteristics that their R intensity value is greater than G value and G intensity value is greater than the B. So, for a pixel at spatial location (x, y) to be fire pixel the below rule must be satisfied.

$$R_1(x,y) = \begin{cases} 1, & \text{if } R(x,y) > G(x,y) > B(x,y) \\ 0, & \text{otherwise} \end{cases}$$

------ (1)

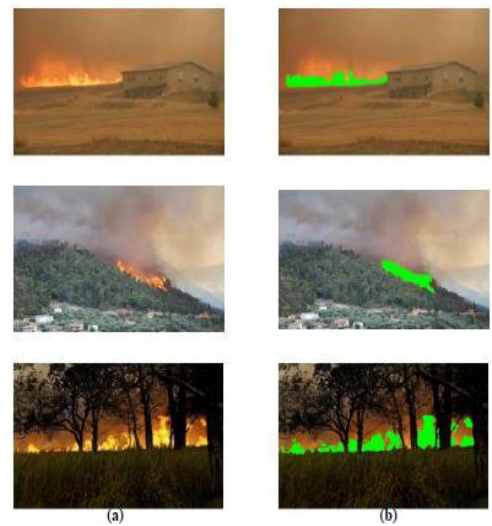

**Fig.4: Original RGB images [19] are given in column (a), and corresponding fire region manually segmented in column (b).**

TABLE I
Mean values of R, G, and B planes of fire regions for images given in Fig.4.

| Row | Mean R | Mean G | Mean B |
|---|---|---|---|
| 1 | 252.5 | 210.3 | 83.1 |
| 2 | 174.4 | 121.1 | 89.4 |
| 3 | 250.1 | 208.6 | 40.8 |

### B. Rule II

From the histogram analysis of the fire location which is manually segmented as, shown in figure 5. We have identified some threshold values for the pixel to be fire. We have threshold values for all the three planes i.e., R, G, and B plane. We have tested this threshold values for hundred of images from our data set.

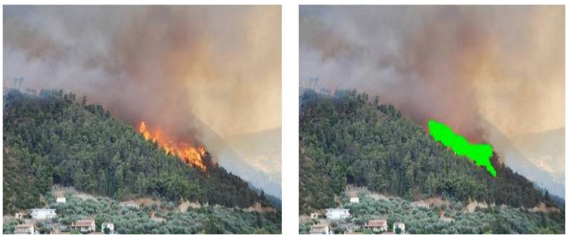

Fig.5: Original RGB image [19] given left and corresponding fire regions, manually labeled with green color are shown in right side.

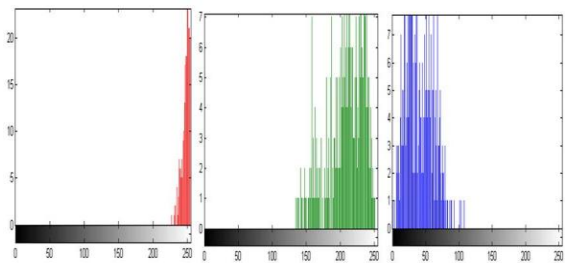

Fig.6: Histogram of manually segmented fire region of image given in Fig.5 R, G and B channel separately.

Based on this observation of a large number of test images, the following rule can be established for detecting a fire pixel at the spatial location (x, y):

$$R_2(x,y) = \begin{cases} 1, & \text{if } (R(x,y) > 190) \cap (G(x,y) > 100) \cap (B(x,y) < 140) \\ 0, & \text{otherwise} \end{cases}$$

------- (2)

When the image is converted from RGB to $YC_bC_r$ color space, intensity and chrominance is easily discriminated. This help to model the fire region easily in $YC_bC_r$ color space.

$$\begin{bmatrix} Y \\ Cb \\ Cr \end{bmatrix} = \begin{bmatrix} 0.2568 & 0.5041 & 0.0979 \\ -0.1482 & -0.2910 & 0.4392 \\ 0.4392 & -0.3678 & -0.0714 \end{bmatrix} \begin{bmatrix} R \\ G \\ B \end{bmatrix} + \begin{bmatrix} 16 \\ 128 \\ 128 \end{bmatrix}$$

------ (a)

where Y is luminance, $C_b$ and $C_r$ are ChrominanceBlue and ChrominanceRed components, respectively.

Given a RGB-represented image, it is converted into $YC_bC_r$ represented color image using the standard RGB-to-$YC_bC_r$.

The mean values of the three components Y, Cb, and Cr, denoted by $Y_{mean}$, $Cb_{mean}$ and $Cr_{mean}$ respectively are computed as follows:

$$Y_{mean}(x,y) = \frac{1}{M \times N} \sum_{x=1}^{M} \sum_{y=1}^{N} Y(x,y)$$

$$Cb_{mean}(x,y) = \frac{1}{M \times N} \sum_{x=1}^{M} \sum_{y=1}^{N} Cb(x,y)$$

$$Cr_{mean}(x,y) = \frac{1}{M \times N} \sum_{x=1}^{M} \sum_{y=1}^{N} Cr(x,y)$$

----- (b)

where, (x,y) denotes the spatial location of pixels, M × N is the total number of pixels in the given image.

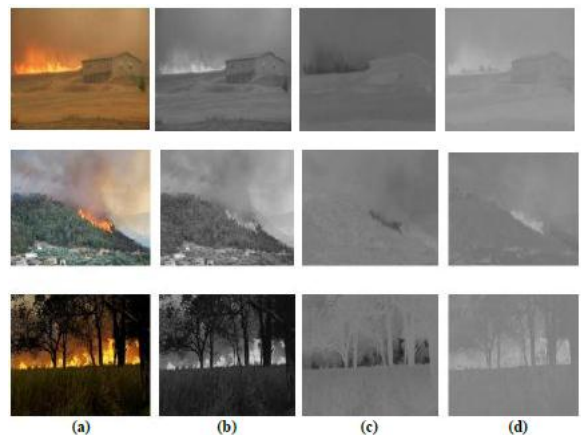

Fig.7: Original RGB image [19] in column [19](a), and Y, $C_b$, and $C_r$ channels in column (b)-(d), respectively.

TABLE II
Mean values of Y, $C_b$, and $C_r$ planes of fire regions for images given in Fig.7.

| Row index in Fig.7 | Mean Y | Mean $C_b$ | Mean $C_r$ |
|---|---|---|---|
| 1 | 130 | 106 | 153 |
| 2 | 189.3 | 48.1 | 158.2 |
| 3 | 195 | 65.8 | 155.6 |

*C. Rule III and IV*

From the Table II we have developed the following two rules. The rules were observed for a large amount of test images from our data base.

The two rules for detecting a fire pixel at spatial location (x,y):

$$R_3(x,y) = \begin{cases} 1, & \text{if } Y(x,y) \geq Cb(x,y) \\ 0, & \text{otherwise} \end{cases} \quad \text{----- (3)}$$

$$R_4(x,y) = \begin{cases} 1, & \text{if } Cr(x,y) \geq Cb(x,y) \\ 0, & \text{otherwise} \end{cases} \quad \text{------ (4)}$$

*D. Rule V*

The flame region is generally the brightest region in the observed scene, the mean values of the three channels, in the overall image $Y_{mean}$, $Cb_{mean}$, and $Cr_{mean}$ contain valuable information. From the figure 7 it can be observed that for the flame region the value of the Y component is bigger than the mean Y component of the overall image while the value of $C_b$ component is in general smaller than the mean $C_b$ value of the overall image. Furthermore, the Cr component of the flame region is bigger than the mean Cr component.

These observations which are verified over countless experiments with images containing fire regions are formulated as the following rule:

$$R_5(x,y) = \begin{cases} 1, & \text{if } (Y(x,y) \geq Y_{mean}(x,y)) \cap (Cb(x,y) \leq Cb_{mean}(x,y)) \\ & \cap (Cr(x,y) \geq Cr_{mean}(x,y)) \\ 0, & \text{otherwise} \end{cases} \quad \text{-------------- (5)}$$

where, $R_5(x,y)$ indicates that any pixel that satisfies the condition given in eq.(5) is considered to be fire pixel.

*E. Rule VI*

It can easily be observed from Figure 7(c) and Figure 7(d) that there is a significant difference between the Cb and Cr components of the fire pixels. For the fire pixel the Cb component is predominantly "black" (lower intensity) while the Cr component, on the other hand, is predominantly "white" (higher intensity).

This fact can be translated into another rule as follows:

$$R_6(x,y) = \begin{cases} 1, & \text{if } |Cb(x,y) - Cr(x,y)| \geq Th \\ 0, & \text{otherwise} \end{cases} \quad \text{---- (6)}$$

where, the value of Th is accurately determined according to a Receiver Operating Characteristics (ROC) Curve[14].

The ROC curve is obtained by experimenting different values of Th (ranging from 1 to 100) over 100 color images. First, the fire-pixel regions are manually segmented from each color image. The rules 1 through 5 (at a chosen value of Th) are then applied to the manually-segmented fire regions. The true positive is defined as the decision when an image contains a fire, and false positive is defined as the decision when an image contains no fire but classified as having fire. The ROC curve consists of 100 data points corresponding to different Th values, the corresponding true positive (i.e., correct-detection) and false positive (i.e., false-alarm) rates are computed and recorded.

The correct detection is defined as the same decision when an image indeed contains a fire incident, while the false alarm is defined as incorrect decision when an image contains no fire but mis-detected as having fire. For good fire detection system it should not miss any fire alarm, with the established ROC curve, the value of Th is picked such that the systems true positive rate is high enough.

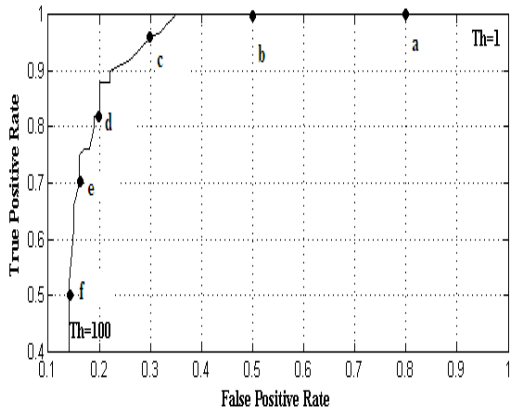

Fig.8: The receiver operating characteristics (ROC) curve for determining the desired value of Th to be used in eq. 6.

From Figure 8 it is clear that, high true positive rate means high false positive rate. Using this tradeoff, in our experiments the value of Th is picked such that the detection rate is over 95% and false alarm rate is less than 30% (point c) which corresponds to Th = 70.

*F. Rule VII*

From the histogram analysis of the fire location which is manually segmented as, shown in Figure 9(c). We have identified some threshold values for the pixel to be fire. We have threshold values for $C_b$ and $C_r$ planes, we are not considering Y plane because it is luminance component and it depends on illumination conditions. We have tested this threshold values for hundred of images from our data set.

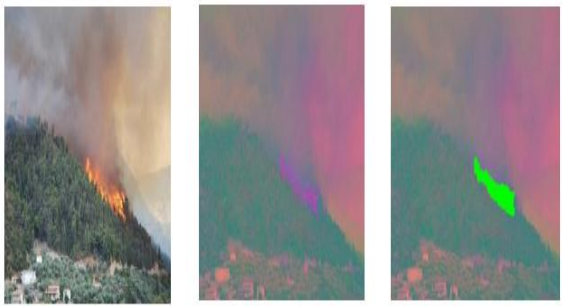

Fig.9: (a) Original RGB image [19], (b) RGB-to-$YC_bC_r$ converted image, and (c) corresponding fire region manually labeled with green color on $YC_bC_r$ image.

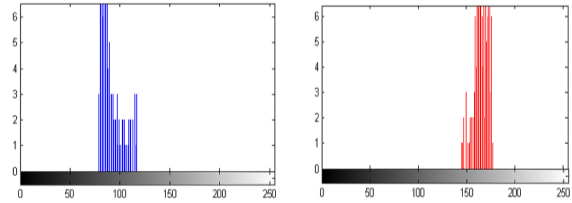

Fig.10: Histogram of manually segmented fire region of image given in fig.9, (a) Cb channel and (b) Cr channel separately.

Based on this observation of a large number of test images, the following rule can be established for detecting a fire pixel at the spatial location (x,y):

$$R_7(x,y) = \begin{cases} 1, & \text{if } (Cb(x,y) \leq 120) \cap (Cr(x,y) \geq 150) \\ 0, & \text{otherwise} \end{cases} \quad \text{--- (7)}$$

A pixel is classified to fire class if all the *Rules I-VII* is satisfied by that pixel. The process of segmentation can be easily understood with the help of figure 11 which explains step by step manner. As can be seen from figure 11 each rule alone will produce false alarm, but their overall combination produces the result in finding out the fire region in the corresponding image. Figure 12 shows the experimental results for different input images.

## IV. PERFORMANCE EVALUATION

For performance evaluation we have used classification error matrix. Table III shows the classification error matrix for the developed classifier. For this we have collected two sets of images from internet. One set composed of images that consist of fire. The fire set consist of 200 images, with diversity in fire-color and environmental illuminations. The other doesn't contain any fire, but contains fire-colored regions such as sun, flowers, reddish objects, etc.

The following condition is used for declaring a fire region: if the model achieves to detect at least 10 pixels as fire, then it is assumed that the image has fire region in it. For false alarm rate the same criterion is used with the non-fire image set.

Classification error matrix is the relationship between known reference data and the corresponding classified result. In this matrix the diagonal elements represents the correctly classified result and non-diagonal elements gives the number of misclassified result.

From the Table IV it can be noted that kappa coefficient is greater than .75 then, the classifier is good[17].

**Table III**
**Classification Error Matrix for the Proposed Classifier.**

|  | Actual data | |  |
|---|---|---|---|
| Classified data | Fire | No Fire |  |
| Fire | 198(A) | 28(B) | 226 (R1) |
| No Fire | 2(C) | 172(D) | 174 (R2) |
|  | 200 (C1) | 200 (C2) |  |

**TABLE IV**
**Different parameters calculated for performance evaluation.**

| Class | Omission Error (%) | Commission Error (%) | Accuracy (%) | | | Kappa Coefficient |
|---|---|---|---|---|---|---|
|  |  |  | UA | PA | OA |  |
| Fire | 1 | 12.38 | 87.6 | 99 | 92.5 | 0.85 |
| No-Fire | 14 | 1.14 | 98.8 | 86 |  |  |

UA- User Accuracy, PA- Producer Accuracy, OA- Overall Accuracy

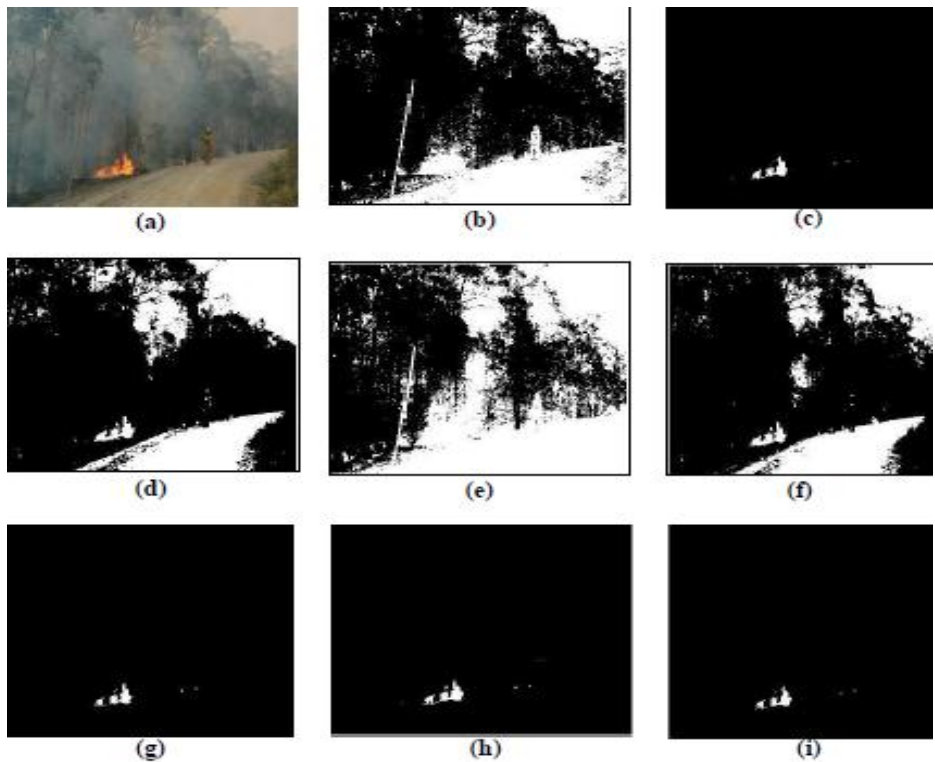

**Fig.11:** Fire detection in still image; (a) original RGB image[19], (b) fire segmented using only rule I, (c) fire segmented using only rule II, (d) fire segmented using only rule III, (e) fire segmented using only rule IV, (f) fire segmented using only rule V, (g) fire segmented using only rule VI, (h) fire segmented using only rule VII, (i) fire region segmented by combining all the rules I- VII.

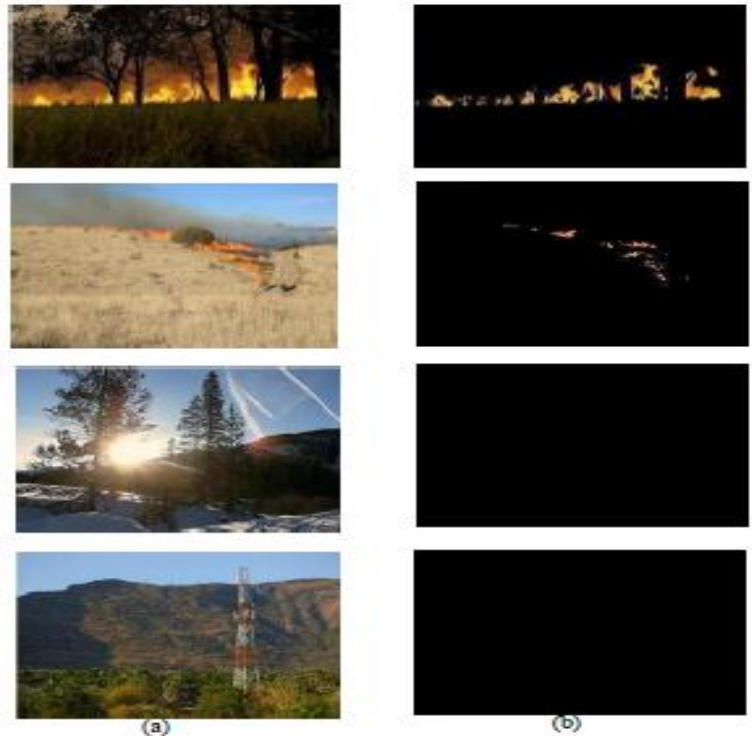

**Fig.12: Experimental Results (a) Input images, (b) corresponding output images.**

## V. CONCLUSION

In this research work a rule based color model for forest fire pixel classification is proposed. The proposed color model makes use of RGB color space and $YC_bC_r$ color space. From this a set of seven rules were defined for the pixels to be classified as fire pixel.

The performance of the proposed algorithm is tested on two sets of images; one containing fire and the other with no- fire images. The proposed model achieves 99% flame detection rate and 14% false alarm rate.

The arithmetic operations of this model are linear with the image size. Also, the algorithm is cheap in computational complexity. This makes it suitable to use in real time forest fire monitoring system.

## VI. FUTURE ENHANCEMENT

The proposed system can be realized in future and can evaluate the performance of the system in real time forest fire monitoring system.

Also, instead of using camera images if we go for videos, then we can calculate the spread of fire with time. Further, the flicker nature of fire can be utilized so as to reduce false alarm rate.